\pdfoutput=1
\documentclass[letterpaper,10pt,journal,twoside]{IEEEtran}
\usepackage{amsmath,amsfonts}
\usepackage{algorithmic}
\usepackage{algorithm}
\usepackage{array}
\usepackage[caption=false,font=normalsize]{subfig}
\usepackage{textcomp}
\usepackage{stfloats}
\usepackage{url}
\usepackage{verbatim}
\usepackage{graphicx}
\usepackage{cite}
\usepackage{hyperref}
\usepackage{bm}
\usepackage{amsthm}
\usepackage{amssymb}
\usepackage{tabularx}

\usepackage{multirow}
\usepackage{xcolor}
\usepackage{siunitx}
\usepackage{eso-pic}
\AddToShipoutPictureBG*{%
  \AtPageUpperLeft{%
    \put(\LenToUnit{0.5\paperwidth},\LenToUnit{-0.38in}){%
      \makebox[0pt][c]{\parbox{0.86\textwidth}{\centering\scriptsize
        \copyright~2026 IEEE. Personal use of this material is permitted. Permission from IEEE must be obtained for all other uses, in any current or future media, including reprinting/republishing this material for advertising or promotional purposes, creating new collective works, for resale or redistribution to servers or lists, or reuse of any copyrighted component of this work in other works.}}}}}

\begin{document}

\title{GRACE: Gradient-Free Robot Action Generation \\
via Combined Diffusion-MPPI Posterior Mean Estimation}

\author{Leesai Park$^{1,\ast}$, Jiho Hong$^{1,\ast}$, and Sanghyun Kim$^{1,\dagger}$%

\thanks{This research was supported by the National Research Foundation of Korea (NRF) grant funded by the Korea government (MSIT) (No. RS-2026-25491322), by the National Research Council of Science and Technology (NST) grant funded by the Korea government (MSIT) (No. GTL25042-000), and by the Korea Basic Science Institute (National Research Facilities and Equipment Center) grant funded by the Ministry of Science and ICT (No. RS-2025-00564593).}%

\thanks{$^{1}$Department of Mechanical Engineering, Kyung Hee University,
Yongin, Republic of Korea.
{\tt\footnotesize \{leesai2000,jihojihyuk,kim87\}@khu.ac.kr}}%

\thanks{$^{\ast}$Leesai Park and Jiho Hong contributed equally to this work and share first authorship.}%

\thanks{$^{\dagger}$Corresponding author: Sanghyun Kim.}%

}

\maketitle

\begin{abstract}
Diffusion policies generate multimodal robot action sequences from demonstrations, but steering them toward deployment-time constraints typically relies on differentiable guidance costs. This excludes many practical safety constraints, such as binary collision checks, joint limits, and black-box rollout costs that are nondifferentiable. We propose Gradient-free Robot Action generation via Combined diffusion-MPPI posterior mean Estimation (GRACE), which guides a pretrained diffusion policy with Model Predictive Path Integral (MPPI) control using only forward cost evaluations. Building on the common score-ascent structure of diffusion and MPPI, GRACE constructs a cost-conditioned guidance posterior at each reverse step and estimates its mean with a single MPPI update centered at the diffusion reverse mean. For differentiable costs, GRACE recovers conventional gradient guidance under a first-order, matched-covariance approximation. GRACE attains higher success rates than diffusion-based and sampling-based baselines in simulation. {On a real 7-DoF manipulator, GRACE achieves 43/45 successes without obstacle collisions, compared with 17/45 for the unguided prior.} Code and experiment videos are available at \url{https://rcilab.khu.ac.kr/grace/}.
\end{abstract}


\section{INTRODUCTION}

\IEEEPARstart{R}{obot} trajectory generation is a longstanding problem in robotics. An autonomous system must produce trajectories that are dynamically feasible, satisfy task objectives, and remain adaptable to environments that change between deployments. Traditional planners such as RRT~\cite{lavalle2001rapidly} explore the state or configuration space to find feasible paths, while trajectory optimizers such as CHOMP~\cite{ratliff2009chomp} refine an initial trajectory by minimizing costs that encode feasibility, smoothness, and task objectives. However, these methods do not explicitly model the multimodal distribution of feasible trajectories that can arise in manipulation and navigation tasks~\cite{osa2020multimodal}.

Generative learning-based methods address this limitation by modeling distributions over trajectories from demonstration data. In particular, diffusion generative models~\cite{ho2020denoising} can represent high-dimensional, multimodal distributions through an iterative denoising process and have been applied to robot trajectory generation~\cite{ajay2022conditional} and visuomotor policy learning~\cite{chi2025diffusion}. However, sampling from this learned distribution alone does not necessarily ensure that the generated trajectories satisfy additional task objectives or constraints.

 \begin{figure}[t]
    \centering
    \includegraphics[width=\columnwidth]{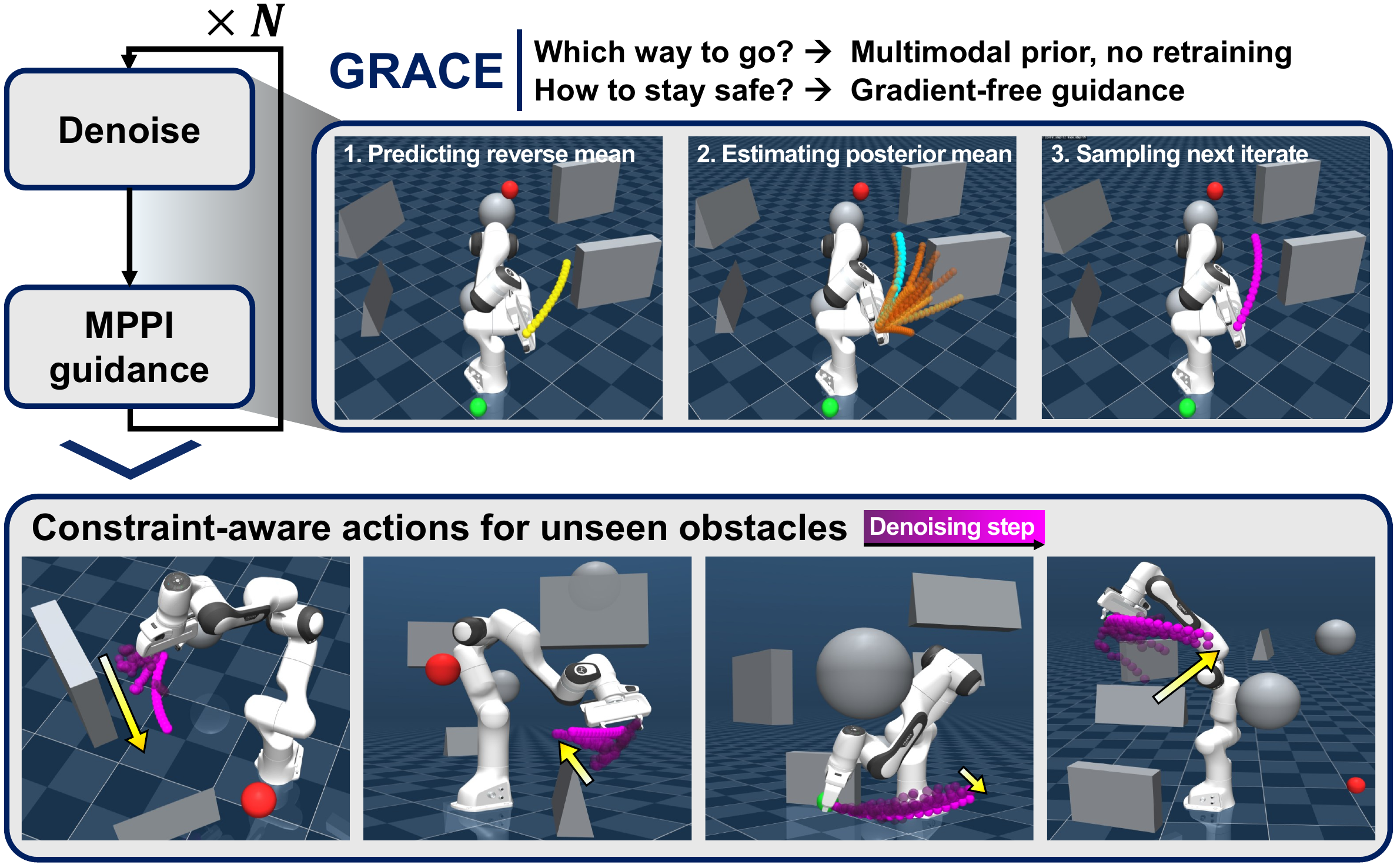}

    \caption{The multimodal diffusion prior decides which way to go without retraining, while gradient-free guidance decides how to stay safe. At each guided denoising step (top), GRACE predicts the diffusion reverse mean (yellow), estimates the cost-conditioned posterior mean (cyan) from MPPI rollouts (orange), and samples the next iterate (magenta). {The bottom panels show different planning scenarios, where trajectories are progressively guided toward constraint satisfaction across denoising steps, visualized from dark to light magenta.} Green and red markers denote the start and goal.}

    \label{fig1}
    \vspace{-5mm}
\end{figure}

Steering the generation process toward such task-specific requirements is the role of guidance. Many deployment-time constraints, such as obstacles that appear only after training, must be imposed at inference time without retraining~\cite{wang2025inference}. Existing gradient-based inference-time methods steer the reverse diffusion process using gradients of task-specific costs evaluated on intermediate denoising samples. MPD~\cite{carvalho2025motion} uses gradients of differentiable motion-planning costs to bias samples toward low-cost trajectories, while EDMP~\cite{saha2024edmp} runs parallel guided denoising processes using an ensemble of collision-cost formulations and guidance schedules. DynaGuide~\cite{du2025dynaguide} instead derives the guidance gradient from a latent outcome-matching objective evaluated through an external learned dynamics model.

However, these approaches require differentiable guidance objectives and may converge to poor local solutions when the resulting optimization landscape is nonconvex~\cite{luo2024potential}. Gradient-free methods remove the differentiability requirement by generating candidate trajectories and scoring them through forward cost evaluations. Diffusion-ES~\cite{yang2024diffusion} uses an outer evolutionary loop that evaluates clean trajectories with a black-box reward and mutates high-scoring samples through truncated diffusion.
{JM2D~\cite{jung2025joint} uses Monte Carlo importance sampling to estimate an interaction-weighted joint score over a diffusion plan and a model-based optimization variable. More closely related to our in-loop setting, D-Cubed~\cite{yamada2024d} applies CEM-inspired guidance within reverse diffusion by selecting the lowest-cost latent trajectory for subsequent denoising.}

We present Gradient-free Robot Action generation via Combined diffusion-MPPI posterior mean Estimation (GRACE), {which formulates each gradient-free reverse-step update as posterior mean estimation over the action trajectory.}
Motivated by the score-ascent correspondence between diffusion and MPPI~\cite{li2025unifying,xue2025full}, {GRACE constructs a Boltzmann cost-conditioned posterior around each diffusion reverse mean and estimates its first moment from MPPI-weighted rollouts.}
{Unlike the single-candidate update in D-Cubed, a forward-KL projection of this posterior onto the Gaussian reverse-kernel family yields the posterior first moment as the guided reverse mean.}
{The resulting reverse kernel is then sampled to obtain the next iterate, as illustrated in Fig.~\ref{fig1}.}
{For differentiable costs, the same formulation recovers conventional one-step gradient guidance under a first-order, matched-covariance approximation.}

The contributions of this paper are as follows:
\begin{itemize}
\item {\textbf{MPPI-guided posterior-mean diffusion.}}
{GRACE integrates MPPI into reverse diffusion by projecting a Boltzmann cost-conditioned posterior onto the Gaussian reverse-kernel family and estimating its mean from MPPI-weighted rollouts, enabling gradient-free adaptation without retraining.}

\item {\textbf{Connection to gradient guidance.}}
{We show that GRACE recovers conventional one-step gradient guidance for differentiable costs under a first-order approximation with the guidance covariance matched to the diffusion reverse covariance.}

\item {\textbf{Experimental validation.}}
{GRACE achieves the highest success rates in 2D and 3D simulations, with analyses of trajectory multimodality, guidance sensitivity, and execution horizons, and achieves 43/45 successes without obstacle collisions across three real-robot obstacle conditions.}
\end{itemize}

\vspace{-1mm}

\section{PRELIMINARIES}
\label{sec:prelim}

\subsection{Diffusion Models for Trajectory Generation}
\label{subsec:diffusion}
We adopt the Denoising Diffusion Probabilistic Model (DDPM) formulation~\cite{ho2020denoising} for trajectory generation.
Let $\mathbf{U}^{(0)} \in \mathbb{R}^{H \times m}$ denote an action sequence of horizon $H$ drawn from a demonstration distribution $p_{\mathrm{data}}$. The forward diffusion process gradually corrupts $\mathbf{U}^{(0)}$ by Gaussian noise over $N$ steps, with $\alpha^{(i)} = 1-\beta^{(i)}$ for a predefined noise schedule $\{\beta^{(i)}\}_{i=1}^{N}$. The $i$-step forward noising distribution can be written in closed form as
\begin{equation}
q(\mathbf{U}^{(i)} \mid \mathbf{U}^{(0)})
= \mathcal{N}\!\left(\mathbf{U}^{(i)};\sqrt{\bar\alpha^{(i)}}\,\mathbf{U}^{(0)},\,(1-\bar\alpha^{(i)})\mathbf{I}\right),
\label{eq:forward_transition}
\end{equation}
where $\bar\alpha^{(i)} = \prod_{j=1}^{i}\alpha^{(j)}$. Equivalently, it can be reparameterized as
\begin{equation}
\mathbf{U}^{(i)} = \sqrt{\bar\alpha^{(i)}}\,\mathbf{U}^{(0)} + \sqrt{1-\bar\alpha^{(i)}}\,\bm{\epsilon},
\qquad \bm{\epsilon}\sim\mathcal{N}(\mathbf{0},\mathbf{I}).
\label{eq:reparam}
\end{equation}
Generation reverses this process through a learned Gaussian kernel
\begin{equation}
p_\theta(\mathbf{U}^{(i-1)} \mid \mathbf{U}^{(i)})
= \mathcal{N}\!\left(\mathbf{U}^{(i-1)};\,\bm{\mu}_\theta(\mathbf{U}^{(i)},i),\,\bm{\Sigma}^{(i)}\right),
\label{eq:reverse_kernel}
\end{equation}
whose mean is parameterized through a learned noise predictor $\bm{\epsilon}_\theta$ as
\begin{equation}
\bm{\mu}_\theta(\mathbf{U}^{(i)},i)
= \frac{1}{\sqrt{\alpha^{(i)}}}\!\left(\mathbf{U}^{(i)} - \frac{\beta^{(i)}}{\sqrt{1-\bar\alpha^{(i)}}}\,\bm{\epsilon}_\theta(\mathbf{U}^{(i)},i)\right),
\label{eq:mu_theta}
\end{equation}
trained by minimizing the denoising objective
\begin{equation}
\mathcal{L}_{\mathrm{diff}}(\theta)
= \mathbb{E}_{i,\mathbf{U}^{(0)},\bm{\epsilon}}\!\left[\,\big\lVert \bm{\epsilon} - \bm{\epsilon}_\theta(\mathbf{U}^{(i)},i)\big\rVert_2^2\,\right].
\label{eq:diff_loss}
\end{equation}

\subsection{Model Predictive Path Integral Control}
\label{subsec:mppi}
Consider a discrete-time system $\mathbf{x}_{t+1} = f(\mathbf{x}_t,\mathbf{v}_t)$ with state $\mathbf{x}_t\in\mathbb{R}^n$ and applied control $\mathbf{v}_t\in\mathbb{R}^m$. Let $\mathbf{V} = (\mathbf{v}_0, \ldots, \mathbf{v}_{H-1})$ denote an applied control sequence. Given $\mathbf{x}_0$, the rollout cost is
\begin{equation}
J(\mathbf{V})
= \sum_{t=0}^{H-1}\ell(\mathbf{x}_t,\mathbf{v}_t) + \phi(\mathbf{x}_H).
\label{eq:rollout_cost}
\end{equation}
The information-theoretic formulation~\cite{williams2018information} yields the optimal control distribution
\begin{equation}
q^\star(\mathbf{V}) = \frac{1}{\eta} \mathrm{e}^{-\frac{1}{\lambda}J(\mathbf{V})} \mathcal{N}(\mathbf{V};\widetilde{\mathbf{U}},\bm{\Sigma}),
\label{eq:mppi_optimal}
\end{equation}
where $\lambda>0$ is a temperature, $\mathcal{N}(\mathbf{V};\widetilde{\mathbf{U}},\bm{\Sigma})$ is a Gaussian base measure centered at $\widetilde{\mathbf{U}}$, and $\eta$ is the normalizer.
The common uncontrolled choice is $\widetilde{\mathbf{U}} = \mathbf{0}$.
MPPI approximates $q^\star$ within the Gaussian family $\mathcal{N}(\mathbf{V};\mathbf{U},\bm{\Sigma})$, optimizing only the mean $\mathbf{U}$ while holding the covariance $\bm{\Sigma}$ fixed, by minimizing the forward-KL divergence. For fixed $\bm{\Sigma}$, this gives the moment-matching solution $\mathbf{U}^\star = \mathbb{E}_{q^\star}[\mathbf{V}]$.

Substituting~\eqref{eq:mppi_optimal} into the moment-matching solution yields a
normalized weighted expectation under the Gaussian base measure.
We then rewrite this expectation under the proposal
$\mathcal{N}(\mathbf{V};\mathbf{U},\bm{\Sigma})$ using importance
sampling. Since the base and proposal distributions share the covariance
$\bm{\Sigma}$, their density ratio can be absorbed into the cost, giving
\begin{equation}
\mathbf{U}^\star
= \frac{\mathbb{E}_{\mathcal{N}(\mathbf{U},\bm{\Sigma})}\!\left[
\mathbf{V}\,\mathrm{e}^{-\frac{1}{\lambda}\widetilde{J}(\mathbf{V})}
\right]}
{\mathbb{E}_{\mathcal{N}(\mathbf{U},\bm{\Sigma})}\!\left[
\mathrm{e}^{-\frac{1}{\lambda}\widetilde{J}(\mathbf{V})}
\right]},
\label{eq:mppi_exact}
\end{equation}
where the augmented cost is
\begin{equation}
\widetilde{J}(\mathbf{V})
= J(\mathbf{V}) + \lambda\sum_{t=0}^{H-1}(\mathbf{u}_t - \widetilde{\mathbf{u}}_t)^\top \bm{\Sigma}^{-1}\mathbf{v}_t.
\label{eq:mppi_aug_cost}
\end{equation}
Using samples ${}^k\mathbf{V} = \mathbf{U} + {}^k\delta\mathbf{U}$ with ${}^k\delta\mathbf{U} \sim \mathcal{N}(\mathbf{0}, \bm{\Sigma})$, the Monte Carlo estimator of~\eqref{eq:mppi_exact} is
\begin{equation}
\begin{gathered}
\mathbf{U}^\star = \mathbf{U} + \sum_{k=1}^{K} {}^k w \cdot {}^k\delta\mathbf{U},\quad
{}^k w = \frac{\mathrm{e}^{-\frac{1}{\lambda}\widetilde{J}({}^k\mathbf{V})}}
{\sum_{l=1}^{K}\mathrm{e}^{-\frac{1}{\lambda}\widetilde{J}({}^l\mathbf{V})}}.
\end{gathered}
\label{eq:mppi_update}
\end{equation}

\begin{figure*}[t]
    \centering
    \scalebox{1}[0.95]{\includegraphics[width=1.0\textwidth]{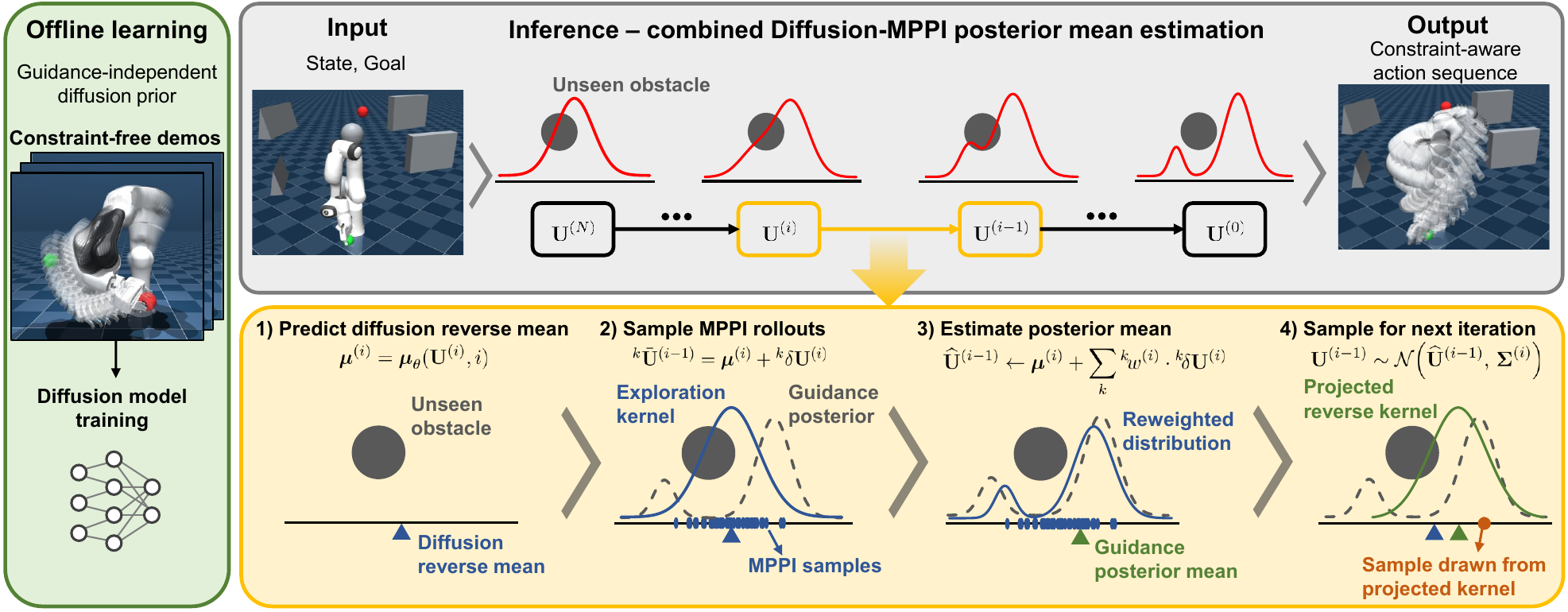}}
    \caption{The GRACE inference pipeline. A diffusion prior learned offline from demonstrations predicts the reverse mean at each denoising step. GRACE samples MPPI rollouts from an exploration kernel centered at this mean and reweights them using deployment-time constraint costs to estimate the guidance posterior mean. The estimated mean defines a projected reverse kernel that retains the diffusion-schedule covariance, from which the next iterate is sampled, enabling constraint-aware action generation without differentiating the cost or the dynamics.}
    \label{systemOverview}
    \vspace{-3mm}
\end{figure*}

\section{PROPOSED METHOD}
\label{sec:method}
In this section, we develop GRACE from the shared score-ascent structure of diffusion and MPPI, which motivates a cost-conditioned posterior whose mean can be estimated through gradient-free MPPI rollouts. The overall inference pipeline is illustrated in Fig.~\ref{systemOverview}.

\subsection{Diffusion and MPPI as Score Ascent}
\label{subsec:score_ascent}
Differentiating the log of~\eqref{eq:forward_transition} yields
\begin{equation}
\nabla_{\mathbf{U}^{(i)}}\log q(\mathbf{U}^{(i)}\mid\mathbf{U}^{(0)})
= {-\bm{\epsilon} \over \sqrt{1-\bar\alpha^{(i)}}},
\label{eq:cond_score}
\end{equation}
which shows that the noise target $\bm{\epsilon}$ is proportional to the score of the forward conditional distribution. At the optimum of the denoising objective~\eqref{eq:diff_loss}, the noise predictor recovers the conditional expectation of $\bm{\epsilon}$ given $\mathbf{U}^{(i)}$. The resulting learned score is
\begin{equation}
\bm{s}_\theta(\mathbf{U}^{(i)},i)
= -\frac{\bm{\epsilon}_\theta^\star(\mathbf{U}^{(i)},i)}{\sqrt{1-\bar\alpha^{(i)}}},
\label{eq:opt_score}
\end{equation}
which matches $\nabla_{\mathbf{U}^{(i)}}\log q(\mathbf{U}^{(i)})$, where $q(\mathbf{U}^{(i)})$ denotes the forward marginal induced by the data distribution and the forward noising process. Substituting~\eqref{eq:opt_score} into~\eqref{eq:mu_theta} expresses the reverse mean approximately as a score-ascent step,
\begin{equation}
\bm{\mu}_\theta(\mathbf{U}^{(i)},i)
\approx
\mathbf{U}^{(i)}
+
\beta^{(i)}\,\bm{s}_\theta(\mathbf{U}^{(i)},i).
\label{eq:diff_score_ascent}
\end{equation}

The same view extends to MPPI. Dropping the control prior in~\eqref{eq:mppi_aug_cost} reduces the augmented cost to $J$, so the denominator of~\eqref{eq:mppi_exact} becomes
\begin{equation}
\Psi(\mathbf{U})
:=
\mathbb{E}_{\mathbf{V}\sim\mathcal{N}(\mathbf{U},\bm{\Sigma})}
\!\left[
\mathrm{e}^{-\frac{1}{\lambda}J(\mathbf{V})}
\right].
\label{eq:psi_def}
\end{equation}
Here, $\Psi(\mathbf{U})$ quantifies the expected Boltzmann weight under Gaussian perturbations around $\mathbf{U}$, and its normalized form
$\bar{\Psi}(\mathbf{U}) := \Psi(\mathbf{U})/Z_\Psi$, where
$Z_\Psi := \int \Psi(\mathbf{U})\,d\mathbf{U}$, defines a density over $\mathbf{U}$. Using the Gaussian mean-derivative identity, noting that $\mathrm{e}^{-{{1}\over{\lambda}} J}$ is independent of $\mathbf{U}$,
\begin{equation}
\nabla_\mathbf{U}\Psi(\mathbf{U})
= \bm{\Sigma}^{-1}\mathbb{E}_{\mathcal{N}(\mathbf{U},\bm{\Sigma})}
\left[
(\mathbf{V}-\mathbf{U})\,
\mathrm{e}^{-\frac{1}{\lambda}J(\mathbf{V})}
\right].
\label{eq:grad_psi}
\end{equation}
Since $Z_\Psi$ is constant, $\nabla_\mathbf{U}\log \bar{\Psi} = \nabla_\mathbf{U}\Psi/\Psi$. Substituting $\mathbf{V} = \mathbf{U} + (\mathbf{V}-\mathbf{U})$ into~\eqref{eq:mppi_exact} and using~\eqref{eq:grad_psi} gives
\begin{equation}
\mathbf{U}^\star = \mathbf{U} + \bm{\Sigma}\,\nabla_\mathbf{U}\log \bar{\Psi}(\mathbf{U}).
\label{eq:mppi_score_ascent}
\end{equation}

Thus the MPPI update~\eqref{eq:mppi_score_ascent} and the approximate diffusion reverse step~\eqref{eq:diff_score_ascent} share the same score-ascent form, each advancing the iterate along a scaled score of an underlying distribution, $q(\mathbf{U}^{(i)})$ for diffusion and $\bar{\Psi}$ for MPPI.

\subsection{Cost-Conditioned Guidance Posterior}
\label{subsec:posterior}
The shared score-ascent structure of Section~\ref{subsec:score_ascent} suggests that steering the reverse process toward low-cost trajectories amounts to combining the two ascent directions, which corresponds to targeting the product of the diffusion reverse kernel and a cost-induced likelihood at each step. Bayes' rule provides exactly this product when the deployment-time objective is represented as a conditioning event $\mathcal{C}$ whose likelihood decreases with the rollout cost $J$. The event $\mathcal{C}$ depends only on the action sequence $\mathbf{U}^{(i-1)}$ and is conditionally independent of $\mathbf{U}^{(i)}$ given $\mathbf{U}^{(i-1)}$. Consequently, applying Bayes' rule to the reverse step gives
\begin{equation}
p(\mathbf{U}^{(i-1)}\!\mid\!\mathbf{U}^{(i)},\mathcal{C})
\!\propto\! p(\mathcal{C}\mid\mathbf{U}^{(i-1)})\,p_\theta(\mathbf{U}^{(i-1)}\mid\mathbf{U}^{(i)}).
\label{eq:posterior_prop}
\end{equation}

A Boltzmann likelihood $p(\mathcal{C}\mid\mathbf{U}^{(i-1)})\propto \mathrm{e}^{-{1\over\lambda}J(\mathbf{U}^{(i-1)})}$ is adopted, consistent with the cost-induced target of Section~\ref{subsec:score_ascent}. For compactness, let $\bm{\mu}^{(i)}:=\bm{\mu}_\theta(\mathbf{U}^{(i)},i)$ and $p_\theta^{(i)}(\mathbf{U}^{(i-1)}):=p_\theta(\mathbf{U}^{(i-1)}\mid\mathbf{U}^{(i)})$ denote the diffusion reverse mean and the diffusion reverse kernel at step $i$, respectively. Substituting the diffusion reverse kernel~\eqref{eq:reverse_kernel} into~\eqref{eq:posterior_prop} gives the reverse-kernel posterior
\begin{equation}
\!\!\!\pi^{(i)}(\mathbf{U}^{(i-1)}\!\!\mid\!\!\mathbf{U}^{(i)}\!,\mathcal{C})
\!\!=\!\!
{\mathrm{e}^{-{1\over\lambda}J(\mathbf{U}^{(i-1)})}\,
\!\!\mathcal{N}\!(\mathbf{U}^{(i-1)};\!\bm{\mu}^{(i)}\!,\!\bm{\Sigma}^{(i)}) \over Z^{(i)}},
\label{eq:posterior_nominal}
\end{equation}
where $Z^{(i)}$ is the normalizing constant.
The reverse-kernel posterior shares the product form of the MPPI optimal distribution~\eqref{eq:mppi_optimal}, with the Gaussian centered at $\bm{\mu}^{(i)}$, so its mean can be estimated from cost-weighted rollouts sampled from $\mathcal{N}(\bm{\mu}^{(i)},\bm{\Sigma}^{(i)})$. However, as the schedule drives $\bm{\Sigma}^{(i)}\!\to\!\mathbf{0}$ for $i\!\to\!1$, the rollouts concentrate near $\bm{\mu}^{(i)}$, restricting the range over which cost weighting can shift the posterior mean.

Accordingly, an exploration kernel centered at the diffusion reverse mean and equipped with an independently controlled covariance is introduced:
\begin{equation}
r_g^{(i)}(\mathbf{U}^{(i-1)}\!\mid\!\mathbf{U}^{(i)})
:= \mathcal{N}(\mathbf{U}^{(i-1)};\bm{\mu}^{(i)},\bm{\Sigma}_g^{(i)}),
\label{eq:base_measure}
\end{equation}
where $\bm{\Sigma}_g^{(i)}\!\succ\!\mathbf{0}$ is a guidance covariance. When $\bm{\Sigma}_g^{(i)}\!-\!\bm{\Sigma}^{(i)}\!\succeq\!\mathbf{0}$, this exploration kernel is the diffusion reverse kernel convolved with an additional Gaussian exploration kernel,
\begin{equation}
r_g^{(i)}(\mathbf{U}^{(i-1)}\!\!\mid\!\mathbf{U}^{(i)}\!)
\!=\! p_\theta^{(i)}\!(\mathbf{U}^{(i-1)})
\!*\! \mathcal{N}\!\bigl(\mathbf{U}^{(i-1)};\!\mathbf{0},\!\bm{\Sigma}_g^{(i)}\!-\!\bm{\Sigma}^{(i)}\!\bigr),
\label{eq:kernel_convolution}
\end{equation}
which preserves the diffusion reverse mean while inflating only the spread. Applying Bayes' rule to $r_g^{(i)}(\mathbf{U}^{(i-1)}\!\mid\!\mathbf{U}^{(i)})$ gives the exploration-kernel posterior, hereafter referred to as the guidance posterior,
\begin{equation}
\!\!\!\rho^{(i)}(\mathbf{U}^{(i-1)}\!\!\mid\!\!\mathbf{U}^{(i)}\!,\mathcal{C})
\!\!=\!\!
{\mathrm{e}^{-{1\over\lambda}\!J(\mathbf{U}^{(i-1)})}\,
\!\!\mathcal{N}(\mathbf{U}^{(i-1)};\!\bm{\mu}^{(i)}\!,\!\bm{\Sigma}_g^{(i)}) \over Z_g^{(i)}},
\label{eq:posterior_guidance}
\end{equation}
where $Z_g^{(i)}$ is the normalizing constant.
This is the exact cost-conditioned posterior associated with the exploration kernel
$r_g^{(i)}(\mathbf{U}^{(i-1)}\mid\mathbf{U}^{(i)})$. When $\bm{\Sigma}_g^{(i)}=\bm{\Sigma}^{(i)}$, the exploration kernel coincides with the diffusion reverse kernel $p_\theta^{(i)}$ and $\rho^{(i)}$ reduces to $\pi^{(i)}$.

\begin{algorithm}[t]
\caption{GRACE inference procedure.}
\label{alg:grace}
\begin{algorithmic}[1]
\REQUIRE $\bm{\epsilon}_\theta$, $\mathbf{x}_0$, $J$, $\{\bm{\Sigma}^{(i)},\bm{\Sigma}_g^{(i)}\}_{i=1}^{N}$, $K$, $\lambda$, $H_{\mathrm{exec}}$, {$N_g$}
\STATE Sample $\mathbf{U}^{(N)} \sim \mathcal{N}(\mathbf{0},\mathbf{I})$
\FOR{$i = N, N-1, \ldots, 1$}
    \STATE Compute reverse mean $\bm{\mu}^{(i)} = \bm{\mu}_\theta(\mathbf{U}^{(i)}, i)$ from~\eqref{eq:mu_theta}
    \STATE $\widehat{\mathbf{U}}^{(i-1)} \leftarrow \bm{\mu}^{(i)}$
    \IF{{$i \leq N_g$}}
        \FOR{$k = 1, \ldots, K$}
            \STATE Sample ${}^{k}\delta\mathbf{U}^{(i)} \sim \mathcal{N}(\mathbf{0},\bm{\Sigma}_g^{(i)})$
            \STATE Form ${}^{k}\bar{\mathbf{U}}^{(i-1)} = \bm{\mu}^{(i)} + {}^{k}\delta\mathbf{U}^{(i)}$
            \STATE Roll out and evaluate $J({}^{k}\bar{\mathbf{U}}^{(i-1)})$
        \ENDFOR
        \STATE ${}^{k}\!w^{(i)} \leftarrow
        \dfrac{\mathrm{e}^{-\frac{1}{\lambda}J({}^{k}\bar{\mathbf{U}}^{(i-1)})}}
        {\sum_{l=1}^{K}\mathrm{e}^{-\frac{1}{\lambda}J({}^{l}\bar{\mathbf{U}}^{(i-1)})}}$
        \STATE $\widehat{\mathbf{U}}^{(i-1)}
        \leftarrow \bm{\mu}^{(i)} + \sum_{k=1}^{K}
        {}^{k}\!w^{(i)} \cdot {}^{k}\delta\mathbf{U}^{(i)}$
    \ENDIF
    \STATE Sample $\mathbf{U}^{(i-1)}
    \sim \mathcal{N}(\widehat{\mathbf{U}}^{(i-1)},\bm{\Sigma}^{(i)})$
\ENDFOR
\STATE Execute first $H_{\mathrm{exec}}$ actions of $\mathbf{U}^{(0)}$; replan.
\end{algorithmic}
\end{algorithm}

\subsection{KL Projection and MPPI Mean Estimation}
\label{subsec:kl_approx}

Although the guidance posterior $\rho^{(i)}$ is well defined, the nonlinear and binary terms in $J$ preclude closed-form expressions for its normalizer and moments, and direct sampling from $\rho^{(i)}$ is not readily available. Instead of handling $\rho^{(i)}$ in full, it is projected onto a Gaussian reverse kernel whose mean is free and whose covariance is fixed to $\bm{\Sigma}^{(i)}$, which reduces the problem to a single moment that admits Monte Carlo estimation. Fixing the covariance to $\bm{\Sigma}^{(i)}$ ensures that the resulting transition remains in the same schedule-covariance family as the diffusion reverse kernel, while the cost-conditioned information contained in $\rho^{(i)}$ is transferred through the mean.

The corresponding forward-KL projection is
\begin{equation}
\mathbf{U}_g^{\star(i-1)}
:=
\arg\min_{\bm{\nu}}
D_{\mathrm{KL}}\!\left(
\rho^{(i)}
\,\middle\|\,
\mathcal{N}
\bigl(
\mathbf{U}^{(i-1)};
\bm{\nu},
\bm{\Sigma}^{(i)}
\bigr)
\right).
\label{eq:kl_obj}
\end{equation}
Since the covariance of the approximating Gaussian is fixed to $\bm{\Sigma}^{(i)}$, the forward-KL minimizer is obtained by matching the first moment of $\rho^{(i)}$,
\begin{equation}
\mathbf{U}_g^{\star(i-1)}
=
\mathbb{E}_{\rho^{(i)}}
\left[
\mathbf{U}^{(i-1)}
\right].
\label{eq:kl_mean}
\end{equation}
Substituting the guidance posterior~\eqref{eq:posterior_guidance} into~\eqref{eq:kl_mean} gives
\begin{equation}
\!\!\mathbf{U}_g^{\star(i-1)}
\!\!=\!\!
\frac{
\mathbb{E}_{
\mathbf{U}^{(i-1)}
\sim
\mathcal{N}
\left(
\bm{\mu}^{(i)},
\bm{\Sigma}_g^{(i)}
\right)
}\!
\!\left[
\mathbf{U}^{(i-1)}
\mathrm{e}^{-{1\over\lambda}J(\mathbf{U}^{(i-1)})}
\right]
}{
\mathbb{E}_{
\mathbf{U}^{(i-1)}
\sim
\mathcal{N}
\left(
\bm{\mu}^{(i)},
\bm{\Sigma}_g^{(i)}
\right)
}\!
\!\left[
\mathrm{e}^{-{1\over\lambda}J(\mathbf{U}^{(i-1)})}
\right]
}.
\label{eq:mean_ratio}
\end{equation}
Although this first moment remains unavailable in closed form, both expectations are now taken under the Gaussian $\mathcal{N}(\bm{\mu}^{(i)},\bm{\Sigma}_g^{(i)})$, removing the need to sample from $\rho^{(i)}$ or to evaluate $Z_g^{(i)}$. This ratio has the same form as the MPPI update~\eqref{eq:mppi_exact} and is estimated with cost-weighted rollouts. Denoting the resulting Monte Carlo estimate by $\widehat{\mathbf{U}}^{(i-1)}$, consistency of the estimator gives
\begin{equation}
\widehat{\mathbf{U}}^{(i-1)}
\longrightarrow
\mathbf{U}_g^{\star(i-1)}
\qquad
\text{as }
K\longrightarrow\infty.
\label{eq:mean_consistency}
\end{equation}
The estimated mean of the guidance posterior is then used to define the projected guided reverse kernel
\begin{equation}
q_g^{(i)}
\bigl(
\mathbf{U}^{(i-1)}
\mid
\mathbf{U}^{(i)},
\mathcal{C}
\bigr)
:=
\mathcal{N}
\bigl(
\mathbf{U}^{(i-1)};
\widehat{\mathbf{U}}^{(i-1)},
\bm{\Sigma}^{(i)}
\bigr).
\label{eq:guided_kernel}
\end{equation}

{In practice, guidance is applied only during the last $N_g$ denoising steps, where the intermediate trajectories are sufficiently denoised for sampled rollouts to provide informative cost evaluations. This also reduces guidance computation by avoiding rollouts at early steps dominated by diffusion noise, following MPD~\cite{carvalho2025motion}.}
The overall GRACE reverse process based on this construction is summarized in Algorithm~\ref{alg:grace}.

\subsection{Gradient Guidance as a Differentiable Special Case}
\label{subsec:taylor_compare}
To relate GRACE to gradient-based diffusion guidance, we analyze the first-order behavior of~\eqref{eq:mean_ratio} for differentiable costs. Define the deviation from the diffusion reverse mean as $\bm{\delta}:=\mathbf{U}^{(i-1)}-\bm{\mu}^{(i)}$. Under the exploration kernel, the induced distribution of $\bm{\delta}$ is $\mathcal{N}(\mathbf{0},\bm{\Sigma}_g^{(i)})$. Let $\nabla J^{(i)}:=\nabla_{\mathbf{U}}J(\mathbf{U})|_{\bm{\mu}^{(i)}}$. The cost is locally approximated as $J(\bm{\mu}^{(i)}+\bm{\delta})\approx J(\bm{\mu}^{(i)})+\bm{\delta}^{\top}\nabla J^{(i)}$. Substitution into~\eqref{eq:mean_ratio} cancels the common factor $\mathrm{e}^{-\frac{1}{\lambda}J(\bm{\mu}^{(i)})}$ and leaves the perturbation exponent
\begin{equation}
-\frac{1}{2}\bm{\delta}^{\top}
\bigl(\bm{\Sigma}_g^{(i)}\bigr)^{-1}\bm{\delta}
-\frac{1}{\lambda}\bm{\delta}^{\top}\nabla J^{(i)}.
\label{eq:linearized_exponent}
\end{equation}
Completing the square factors the resulting integrand into the Gaussian density $\mathcal{N}(\bm{\delta};-\lambda^{-1}\bm{\Sigma}_g^{(i)}\nabla J^{(i)},\,\bm{\Sigma}_g^{(i)})$ and a constant independent of $\bm{\delta}$. This constant is common to the numerator and the denominator of~\eqref{eq:mean_ratio} and cancels in the ratio, which then reduces to the mean of $\bm{\delta}$ under the shifted Gaussian.

Since $\mathbf{U}^{(i-1)}=\bm{\mu}^{(i)}+\bm{\delta}$, it follows that
\begin{equation}
\mathbf{U}_g^{\star(i-1)}
\approx
\bm{\mu}^{(i)}
-\frac{1}{\lambda}
\bm{\Sigma}_g^{(i)}
\nabla J^{(i)}.
\label{eq:taylor_limit}
\end{equation}
Substituting~\eqref{eq:taylor_limit} into~\eqref{eq:guided_kernel} gives, under the first-order cost approximation and in the infinite-sample limit,
\begin{equation}
\mathbf{U}^{(i-1)}
\sim
\mathcal{N}\!\left(
\bm{\mu}^{(i)}
-\frac{1}{\lambda}
\bm{\Sigma}_g^{(i)}
\nabla J^{(i)},
\bm{\Sigma}^{(i)}
\right).
\label{eq:grad_update}
\end{equation}
When $\bm{\Sigma}_g^{(i)}=\bm{\Sigma}^{(i)}$,~\eqref{eq:grad_update} reduces to the Taylor-linearized one-step gradient-guided update of~\cite{carvalho2025motion}. Thus, gradient guidance is recovered as the differentiable, matched-covariance special case of GRACE.

\begin{table}[t]
\caption{Experimental setup for the two evaluation environments.}
\label{tab:exp_setup}
\centering
\scriptsize
\begin{tabularx}{\columnwidth}{@{}l>{\centering\arraybackslash}X>{\centering\arraybackslash}X@{}}
\hline
 & Point-mass & FR3 \\
\hline
Action                & 2D velocity        & $\Delta\mathbf{q}$ \\
Training demos   & 10{,}000   & 20{,}000 \\
Goal tolerance        & 0.05 m             & 0.03 rad \\
Step budget           & 1{,}000            & 5{,}000 \\
Obstacles at inference& fixed + added      & 6 random \\
Planning horizon $H$  & 16                 & 32 \\
Executed per replan $H_{\mathrm{exec}}$ & 8 & 16 \\
{Guidance samples $K$} & {1{,}000} & {1{,}000} \\
{Guided steps $N_g$} & {10} & {5} \\
\hline
\end{tabularx}
\vspace{-5mm}
\end{table}

\section{EXPERIMENTS}
\label{sec:experiments}
We validate the effectiveness of GRACE on two simulated domains, planar point-mass navigation and 7-DoF Franka Research 3 (FR3) goal-reaching, as well as a real-world experiment with the FR3 manipulator. Complete hyperparameter settings for GRACE and implementation details are available on the project page at \url{https://rcilab.khu.ac.kr/grace/}.

\subsection{Experimental Setup}
\label{subsec:setup}
All experiments were performed on an NVIDIA GeForce RTX 4060 GPU. We use a goal-conditioned Diffusion Policy~\cite{chi2025diffusion} with either a FiLM-conditioned 1D temporal UNet, denoted \textbf{GRACE (UNet)}, or a lightweight FiLM-conditioned 1D residual CNN, denoted \textbf{GRACE (CNN)}.

\begin{figure}[t]
    \vspace{-0.7\baselineskip}
    \centering
    \subfloat[]{\includegraphics[width=0.48\linewidth]{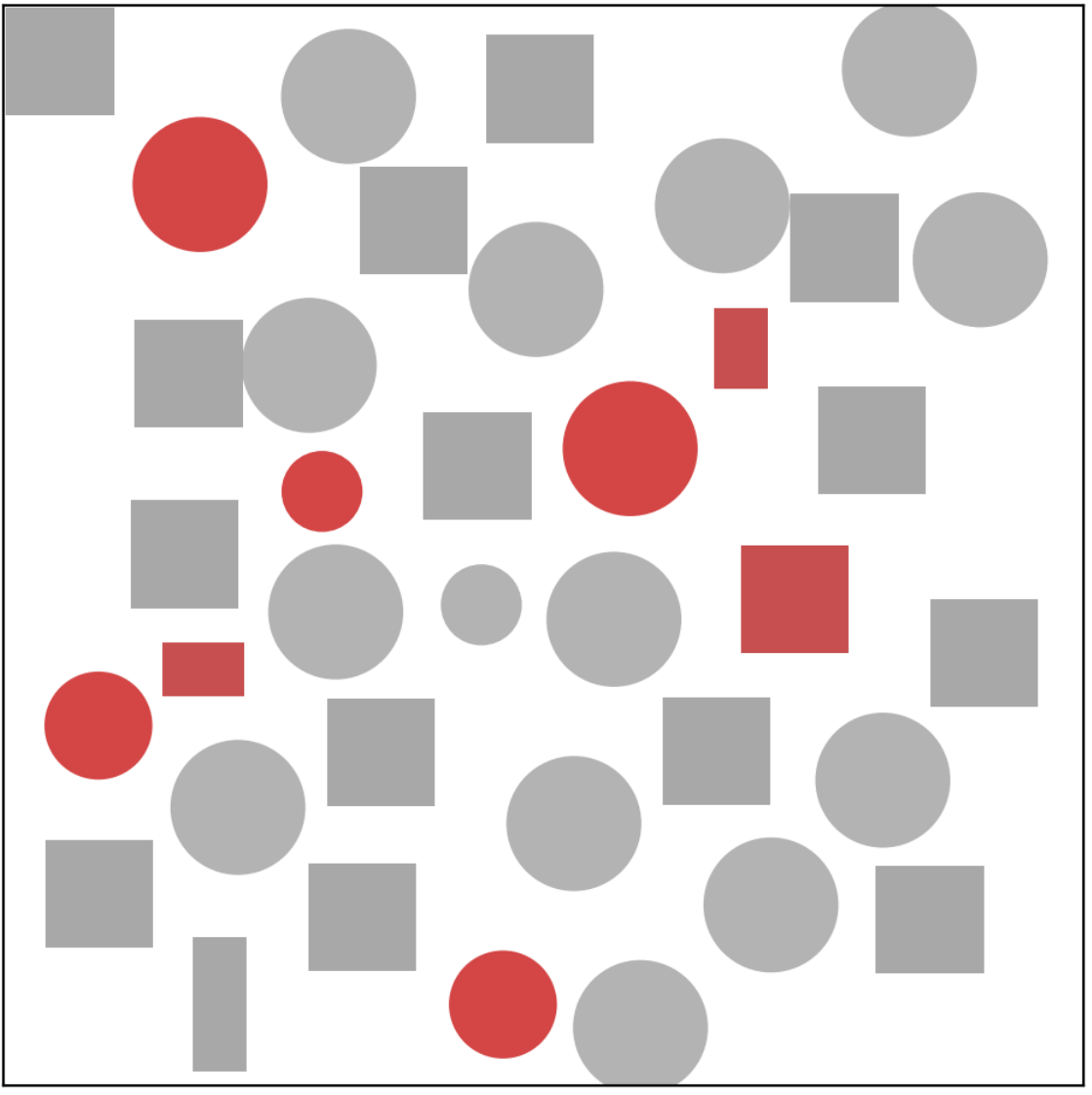} \label{fig3a}}
    \hfill
    \subfloat[]{\includegraphics[width=0.48\linewidth]{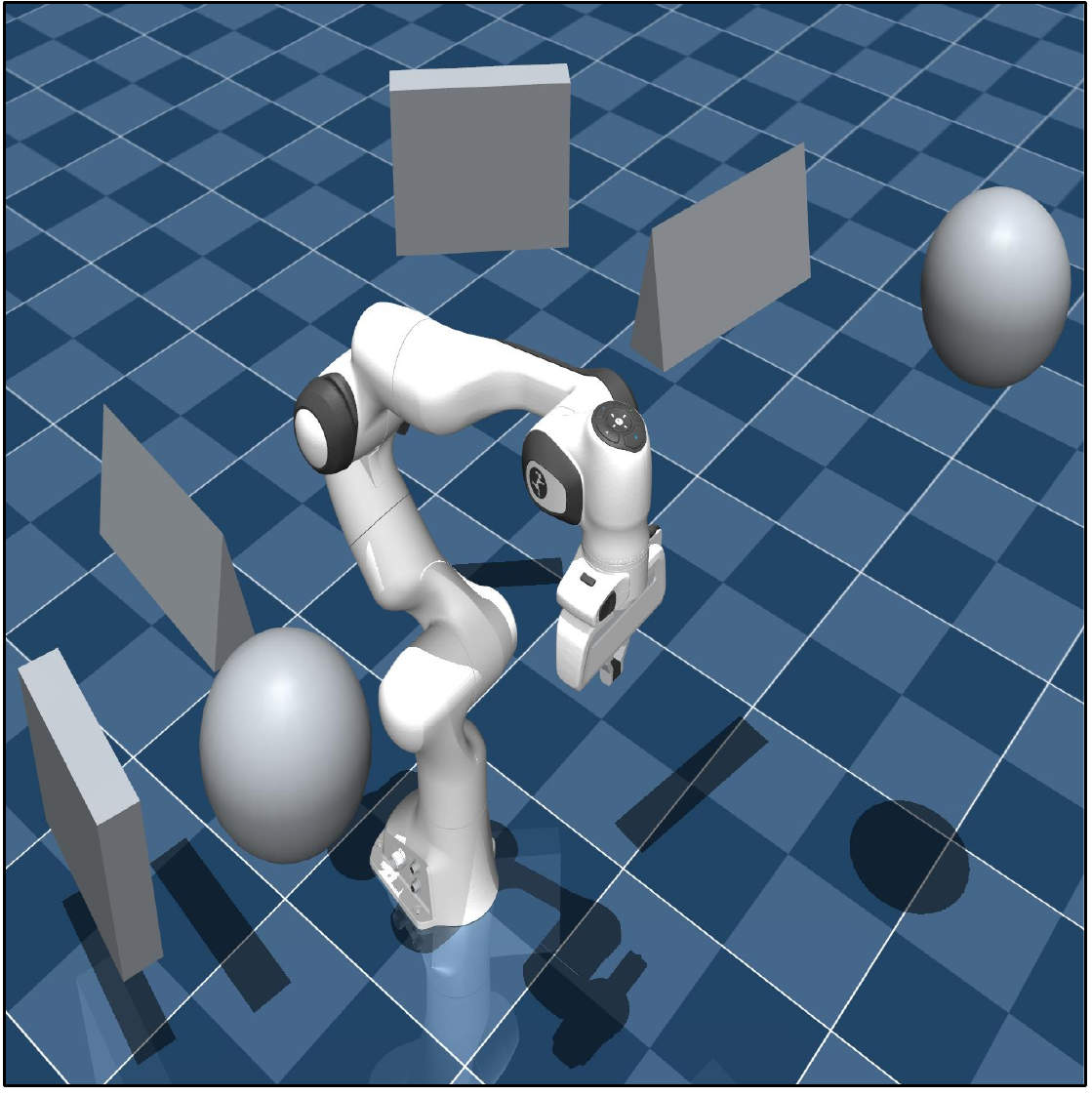} \label{fig3b}}
    \caption{Simulation environments for the 2D and 3D planning tasks. (a) Planar point-mass navigation, where the red obstacles are absent during training and introduced only at inference time. (b) 7-DoF FR3 manipulator goal-reaching with randomly placed obstacles, all of which are introduced only at inference time.} \label{fig3}
    \vspace{-3mm}
\end{figure}

\noindent\textbf{Environments.}
Table~\ref{tab:exp_setup} summarizes the setup of the two evaluation environments.
The first is the planar point-mass navigation environment of MPD~\cite{carvalho2023motion}, shown in Fig.~\ref{fig3a}, where the prior is trained on planner-generated demonstrations collected on a map with fixed obstacles. The start and goal are randomly generated, and additional obstacles absent during training are introduced only at inference time. We conduct {three evaluations: navigation under unseen obstacles over 30 random trials, multimodality across 10 start-goal configurations, and sensitivity to $\lambda$, $\boldsymbol{\Sigma}_g$, $K$, and the guidance update.}  {For the multimodality evaluation, we run each method 20 times per configuration and cluster the successful trajectories after uniform arc-length resampling, treating each cluster as one trajectory mode.}

The second environment evaluates a 7-DoF manipulator reaching goal joint configurations among 3D obstacles, as shown in Fig.~\ref{fig3b}. The prior is trained on obstacle-free feedback-tracking demonstrations, while the obstacles and the start and goal configurations are randomly generated at inference time over 30 trials. {For both environments, simulated time does not advance during planning, and a trial succeeds if the system reaches the target within the task-specific tolerance without timeout or safety violations.}

\noindent\textbf{Cost Formulation.}
Let $d_o(\mathbf{x}_t)$ be the signed distance from state $\mathbf{x}_t$ to obstacle $o$, with penetration depth $p_o(\mathbf{x}_t)=\max(0,-d_o(\mathbf{x}_t))$. We define the obstacle cost over the horizon $H$ as
\begin{equation}
J_{\mathrm{obs}} =
\sum_{t=1}^{H}
\left(
\mathbf{1}_{\mathrm{obs}}(\mathbf{x}_t)
+
\sum_{o \in \mathcal{O}}
\left(
\frac{p_o(\mathbf{x}_t)}{\ell_o}
\right)^2
\right),
\label{eq:cost_obs}
\end{equation}
where $\mathbf{1}_{\mathrm{obs}}$ is the exact binary collision indicator used for constraint checking, $\ell_o$ is a per-obstacle normalization length, and the second term is a continuous soft-penetration penalty that provides a gradient signal where the binary term does not.
For diffusion-guided methods the goal is supplied by the prior, so the total cost is
\begin{equation}
J = w_{\mathrm{obs}} J_{\mathrm{obs}} + w_{\mathrm{prior}} J_{\mathrm{prior}},
\label{eq:cost_guided}
\end{equation}
where $J_{\mathrm{prior}}$ penalizes deviation from the diffusion reverse mean and is defined as
\begin{equation}
J_{\mathrm{prior}}(\mathbf V)
=
\left\|
\mathbf V-\bm{\mu}^{(i)}
\right\|_2^2.
\label{eq:cost_prior}
\end{equation}
Since MPPI estimates a cost-weighted posterior mean, constraint costs alone may shift the guided mean toward safe but goal-ineffective trajectories. The prior term regularizes this shift to preserve goal-directed behavior, while the diffusion-free baselines replace $J_{\mathrm{prior}}$ with an explicit goal cost $J_{\mathrm{goal}}$.

In the 2D environment all methods share the same cost in~\eqref{eq:cost_obs} over the exact geometry, so that the comparison isolates the difference between sampling- and gradient-based optimization under identical conditions. In the 3D environment the gradient-guided methods instead operate on a smoothed approximate geometry, whereas the sampling-based methods evaluate the exact geometry through the binary term alone. The 3D setting further adds self-collision, floor contact, and joint-limit constraints, aggregated as $J_{\mathrm{safe}} = w_{\mathrm{obs}} J_{\mathrm{obs}} + w_{\mathrm{self}} J_{\mathrm{self}} + w_{\mathrm{floor}} J_{\mathrm{floor}} + w_{\mathrm{jl}} J_{\mathrm{jl}}$, which replaces $w_{\mathrm{obs}} J_{\mathrm{obs}}$ in~\eqref{eq:cost_guided}.

\noindent\textbf{Baselines.}
We compare against three families. The first consists of diffusion-free sampling-based planners, namely \textbf{CEM}~\cite{rubinstein1999cross}, \textbf{MPPI}~\cite{williams2018information}, and \textbf{DA-MPPI}, a Diffusion-Annealed MPPI variant inspired by DIAL-MPC~\cite{xue2025full} that anneals the sampling variance across updates.
{The second family consists of gradient-based diffusion methods.} \textbf{Gradient-Guided Diffusion Policy (GG-DP)} applies in-loop refinement during denoising following MPD~\cite{carvalho2025motion}, whereas {\textbf{Post-Optimization Gradient (PO-Grad)} applies gradient refinement after denoising.}
{The third family consists of sampling-based diffusion methods. \textbf{PO-MPPI} applies MPPI refinement after denoising.} \textbf{Diffusion-ES}~\cite{yang2024diffusion} applies gradient-free evolutionary search outside the denoising loop. {\textbf{CEM-DP}, inspired by D-Cubed~\cite{yamada2024d}, applies CEM-based guidance within reverse diffusion.}
All diffusion-based baselines use the same UNet prior as GRACE. {All in-loop and post-optimization diffusion guidance methods use the same $N_g$ guidance updates. All diffusion-free sampling-based planners and sampling-based guidance methods use $K=1{,}000$ samples per update, applying the same control perturbation across the time horizon for each sample, following the single-instance sampling scheme of~\cite{kim2025single} to improve sample efficiency and rollout smoothness.}

\begin{table}[t]
\caption{Quantitative comparison of planning methods in the 2D navigation task.}
\label{tab:2d_mass_results}
\centering
\scriptsize
\begin{tabular*}{\columnwidth}{@{\extracolsep{\fill}}lccc@{}}
\hline
Method
& Success [\%]
& Collision
& Path Length [m]\\
\hline

CEM
& 8/30 (26.7\%)
& 0
& $1.247 \pm 0.143$ \\

MPPI
& 8/30 (26.7\%)
& 0
& $1.241 \pm 0.123$ \\

DA-MPPI
& 8/30 (26.7\%)
& 0
& $1.233 \pm 0.120$ \\

GG-DP
& 18/30 (60.0\%)
& 12
& $2.302 \pm 1.115$ \\

PO-Grad
& 18/30 (60.0\%)
& 12
& $2.657 \pm 1.021$ \\

PO-MPPI
& 22/30 (73.3\%)
& 2
& $2.546 \pm 1.340$ \\

Diffusion-ES
& 7/30 (23.3\%)
& 23
& $2.700 \pm 0.832$ \\

CEM-DP
& 15/30 (50.0\%)
& 15
& $1.882 \pm 0.636$ \\

GRACE (CNN)
& 25/30 (83.3\%)
& 2
& $3.158 \pm 1.379$ \\

GRACE (UNet)
& 25/30 (83.3\%)
& 1
& $2.688 \pm 1.224$ \\

\hline
\end{tabular*}
\vspace{-3mm}
\end{table}

\begin{table}[t]
\caption{Sensitivity analysis of GRACE in the 2D navigation task.}
\label{tab:sensitivity}
\centering
\scriptsize
\begin{tabular*}{\columnwidth}{@{\extracolsep{\fill}}lccc|ccc@{}}
\hline
Update & $\lambda$ & $\boldsymbol{\Sigma}_g$ & $K$
& Success & Collision & Timeout \\
\hline
Weighted & 0.5     & $\operatorname{diag}(0.3^2)$ & 1{,}000 & 25/30 & 1  & 4 \\
Weighted & 0.005   & $\operatorname{diag}(0.3^2)$ & 1{,}000 & 21/30 & 2  & 7 \\
Weighted & 0.00005 & $\operatorname{diag}(0.3^2)$ & 1{,}000 & 22/30 & 7  & 1 \\
Best sample & $-$  & $\operatorname{diag}(0.3^2)$ & 1{,}000 & 21/30 & 9  & 0 \\
\hline
Weighted & 0.5 & $\operatorname{diag}(0.3^2)$ & 300 & 22/30 & 2 & 6 \\
Weighted & 0.5 & $\operatorname{diag}(0.3^2)$ & 50  & 19/30 & 2 & 9 \\
\hline
Weighted & 0.5 & $\operatorname{diag}(0.1^2)$ & 1{,}000 & 17/30 & 13 & 0 \\
Weighted & 0.5 & $\operatorname{diag}(0.7^2)$ & 1{,}000 & 2/30  & 0  & 28 \\
\hline
\end{tabular*}
\vspace{-3mm}
\end{table}

\begin{figure}[t]
    \vspace{-0.7\baselineskip}
    \centering
    \subfloat[]{
        \includegraphics[height=4.4cm, trim=0 0 0 0, clip]{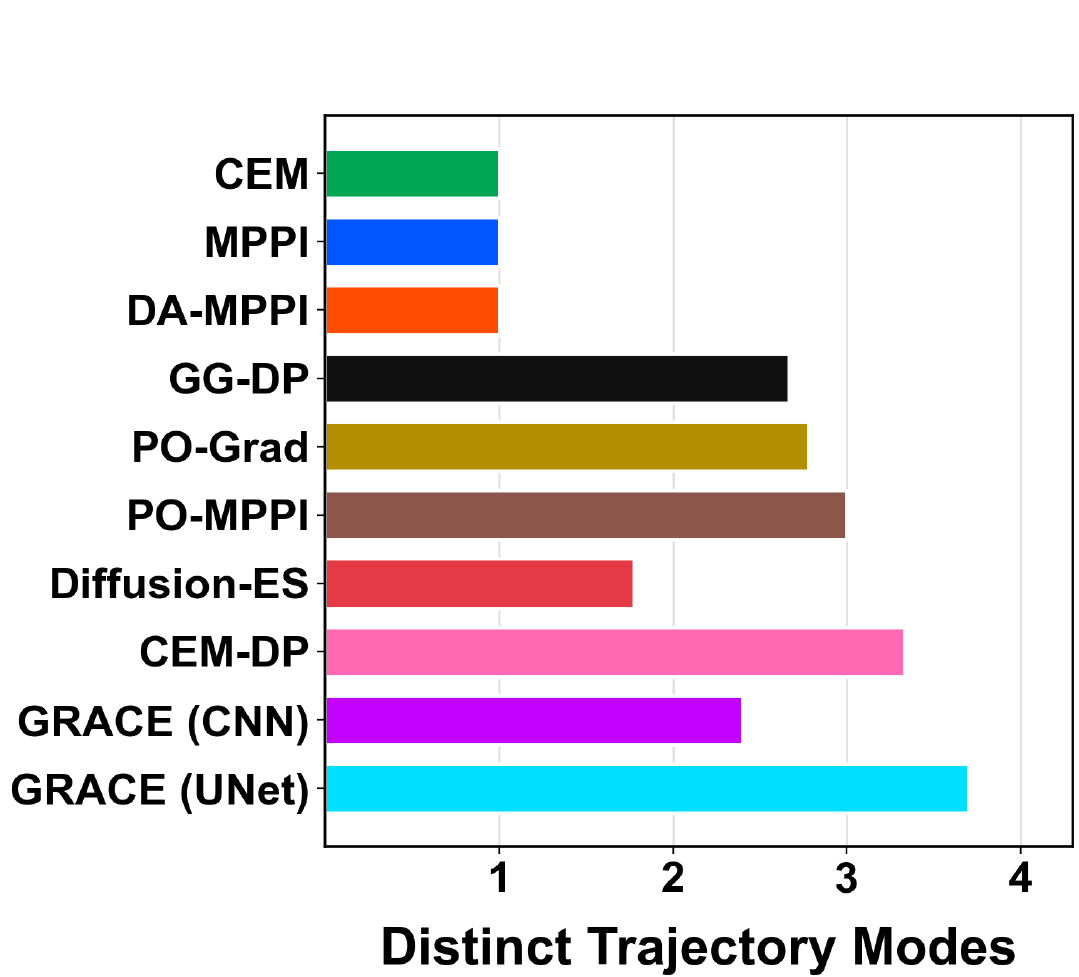}
        \label{fig4a}
    }
    \subfloat[]{
        \includegraphics[height=4.4cm]{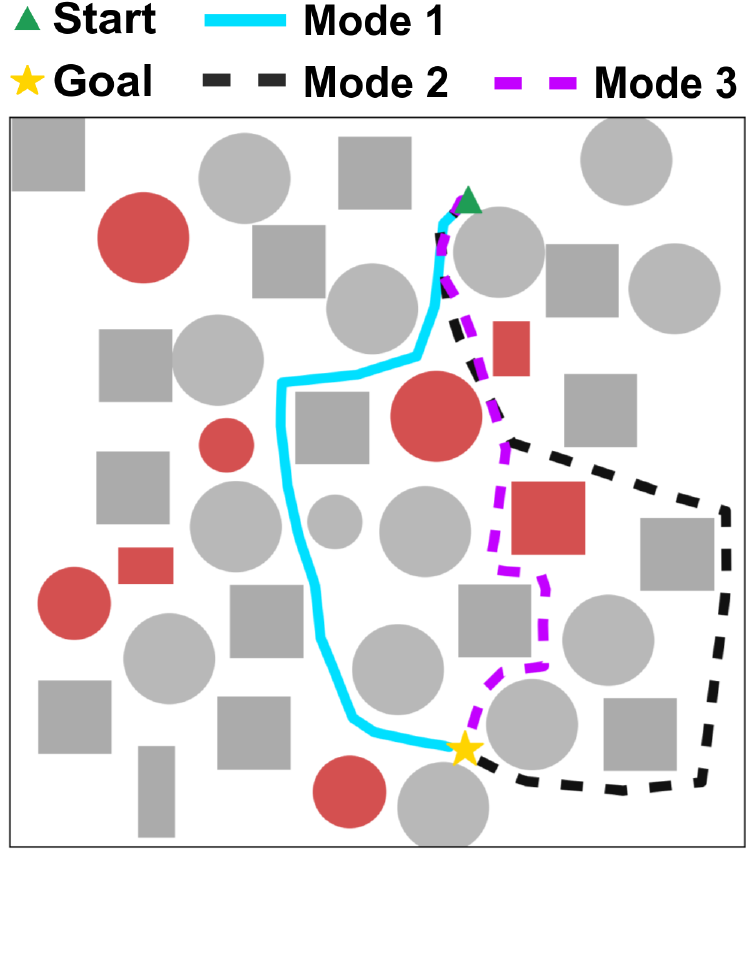}
        \label{fig4b}
    }
    \caption{Multimodality evaluation in the 2D point-mass navigation task. {(a) Average number of trajectory modes produced by each method over the start-goal configurations with at least one successful trajectory, measuring the diversity of generated plans. (b) Trajectory modes produced by GRACE (UNet) for one representative start-goal configuration.}}
    \label{fig4}
    \vspace{-5mm}
\end{figure}

\subsection{Results}
\label{subsec:results}

\noindent\textbf{2D navigation performance.}
Table~\ref{tab:2d_mass_results} summarizes the planar results. The diffusion-free planners CEM, MPPI, and DA-MPPI all reach $26.7\%$ success with no collisions. Although conservative around obstacles, they lack a learned prior, so goal-cost-driven proposals favor short, near-straight routes over feasible detours and often become trapped in local minima. Their success-only path lengths around $1.24$\,m thus reflect easier configurations rather than better planning quality. GG-DP and PO-Grad both reach $60.0\%$ success but incur 12 collision failures, as local gradients can reduce collision cost without directing trajectories toward feasible passages, leading to infeasible refinement in narrow regions.

Among the sampling-based diffusion methods, {PO-MPPI reaches $73.3\%$ success with only two collision failures. Compared with PO-Grad, sampling multiple trajectory perturbations rather than following a local gradient enables broader exploration of alternative detours. However, the final diffusion trajectory remains a fixed prior reference throughout post-optimization, penalizing substantial deviations even when needed to avoid unseen obstacles.} Diffusion-ES reaches only $23.3\%$ success and incurs 23 collision failures because the outer-loop search relies on complete trajectories generated from the learned prior, limiting adaptation to unseen obstacles. {CEM-DP reaches $50.0\%$ success, with all 15 failures due to collision. CEM-DP samples using the diffusion-schedule variance and propagates only the lowest-cost candidate. Decreasing variance restricts exploration, while single-candidate selection discards information from other samples and can steer denoising toward a locally low-cost but collision-prone route.} GRACE achieves the highest success rate of $83.3\%$ with both backbones, with two collision failures for CNN and one for UNet. {The independently controlled guidance covariance maintains exploration, allowing the cost-weighted update to redirect the trajectory even as the diffusion variance decreases. Moreover, because each guided iterate becomes the next input to the denoiser, the prior can redirect the trajectory toward a different demonstrated detour, rather than deforming a fixed trajectory as in PO-MPPI.} GRACE's longer successful paths reflect the detours required to solve harder configurations where other methods fail.

\noindent{\textbf{Trajectory multimodality.}}
{Fig.~\ref{fig4a} reports the average number of distinct trajectory modes over configurations with at least one successful trajectory. The diffusion-free planners recover only one mode, consistent with goal-cost-driven proposals favoring direct routes over alternative feasible detours. Among diffusion-based methods, Diffusion-ES shows the lowest diversity with 1.78 modes, which may reflect outer-loop search over complete trajectories rather than guidance during denoising. GRACE (CNN) and GRACE (UNet) instead recover 2.40 and 3.70 modes, respectively. Although each guided reverse transition uses a Gaussian kernel, the projection is conditioned on the current reverse sample, and the next iterate is sampled with the diffusion-schedule covariance. Different reverse trajectories can therefore evolve toward distinct modes rather than collapsing to a single path.} Fig.~\ref{fig4b} shows the trajectory modes recovered by GRACE (UNet) for one representative start-goal configuration.

\noindent{\textbf{Sensitivity analysis.}}
{Table~\ref{tab:sensitivity} summarizes the 2D sensitivity of GRACE (UNet) to the temperature $\lambda$, guidance samples $K$, and guidance covariance $\boldsymbol{\Sigma}_g$. Decreasing $\lambda$ concentrates the importance weights, reducing the effective sample size from 619 to 62 and finally to 1, while collision failures increase. The best-sample update incurs 9 collision failures and no timeouts. Unlike best-sample selection, which relies on a single lowest-cost candidate, cost-weighted aggregation incorporates multiple low-cost samples into the guided mean, reducing dependence on any single candidate and yielding fewer collisions. Reducing $K$ lowers success, especially at $K=50$, where fewer useful rollouts hinder goal progress and increase timeouts. Guidance covariance shows an exploration-convergence trade-off. $\boldsymbol{\Sigma}_g=\operatorname{diag}(0.1^2)$ restricts exploration and causes frequent collisions, whereas $\boldsymbol{\Sigma}_g=\operatorname{diag}(0.7^2)$ avoids collisions but rarely converges due to overly broad perturbations. Intermediate guidance scales better balance safety and convergence.}

\begin{figure*}[t]
    \centering
    \includegraphics[width=1.0\textwidth]{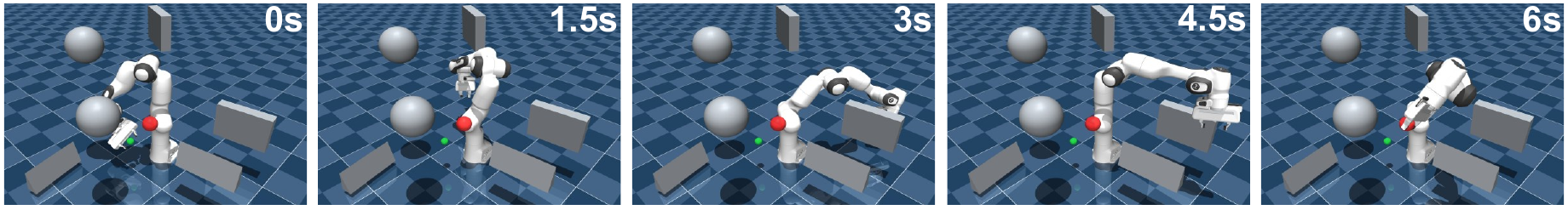}
    \caption{Representative execution snapshots of GRACE in the 3D FR3 manipulator goal-reaching task. The manipulator reaches the target configuration while satisfying multiple safety constraints.}
    \label{fig5}
    \vspace{-3mm}
\end{figure*}

\begin{table*}[t]
\caption{Quantitative comparison of planning methods in the 7-DoF FR3 manipulator goal-reaching task.}
\label{tab:main_results}
\centering
\scriptsize

\begin{tabular*}{\textwidth}{
@{\extracolsep{\fill}}
l
c
ccccc
S[
  table-format = 4.2(4),
  uncertainty-mode = separate
]
S[
  table-format = 3.2(3),
  uncertainty-mode = separate
]
@{}
}
\hline
\multirow{2}{*}{Method}
& \multirow{2}{*}{\shortstack{Success [\%]}}
& \multicolumn{5}{c}{Failure Modes}
& {\multirow{2}{*}{\shortstack{Computation Time [ms]}}}
& {\multirow{2}{*}{\shortstack{Guidance Time [ms]}}} \\
\cline{3-7}
& & Obstacle & Self-collision & Joint-limit & Floor & Timeout & & \\
\hline

CEM
& 24/30 (80.0\%)
& 1 & 0 & 0 & 0 & 5
& 153.45(255)
& \multicolumn{1}{c}{
    $\phantom{244.72}\mathbin{-}\phantom{8.76}$
  } \\

MPPI
& 23/30 (76.7\%)
& 2 & 0 & 5 & 0 & 0
& 29.88(49)
& \multicolumn{1}{c}{
    $\phantom{244.72}\mathbin{-}\phantom{8.76}$
  } \\

DA-MPPI
& 23/30 (76.7\%)
& 6 & 0 & 0 & 0 & 1
& 154.22(249)
& \multicolumn{1}{c}{
    $\phantom{244.72}\mathbin{-}\phantom{8.76}$
  } \\

GG-DP
& {13/30 (43.3\%)}
& {8}
& {2}
& {3}
& {1}
& {3}
& 1059.03(798)
& 644.57(495) \\

{PO-Grad}
& {13/30 (43.3\%)}
& {5}
& {1}
& {6}
& {1}
& {4}
& 956.02(724)
& 546.50(312) \\

{PO-MPPI}
& {26/30 (86.7\%)}
& {3}
& {1}
& {0}
& {0}
& {0}
& 629.35(925)
& 181.45(168) \\

Diffusion-ES
& 12/30 (40.0\%)
& 16 & 0 & 0 & 2 & 0
& 1664.35(672)
& 217.87(70) \\

{CEM-DP}
& {10/30 (33.3\%)}
& {16}
& {1}
& {1}
& {2}
& {0}
& 600.97(1619)
& 181.55(163) \\

GRACE (CNN)
& {27/30 (90.0\%)}
& {2}
& {0}
& {0}
& {0}
& {1}
& 258.14(292)
& 181.66(151) \\

GRACE (UNet)
& {28/30 (93.3\%)}
& {1}
& {0}
& {0}
& {0}
& {1}
& 597.72(787)
& 181.36(118) \\

\hline
\end{tabular*}
\vspace{-4mm}
\end{table*}

\noindent\textbf{3D goal-reaching.}
Unlike the 2D task, the prior here is trained on obstacle-free feedback-tracking demonstrations and converges directly to the goal, so performance depends mainly on how the guidance handles the multiple safety constraints. As shown in Table~\ref{tab:main_results}, the diffusion-free sampling planners perform well in this setting, with CEM reaching $80.0\%$ success and MPPI and DA-MPPI reaching $76.7\%$. Their failures concentrate on joint limits for MPPI under its fixed exploration variance, obstacle collisions for DA-MPPI as annealing reduces late-stage exploration, and near-goal timeouts for CEM.

GG-DP and PO-Grad both reach $43.3\%$ success with failures across several safety constraints, as combining the safety gradients into a single weighted direction can leave conflicting constraints unresolved, at a substantially higher guidance cost than GRACE. {PO-MPPI achieves $86.7\%$ success in this 3D task, suggesting that the limitations of post-update refinement are less pronounced than in 2D navigation, where avoiding unseen obstacles often requires substantial changes to the generated trajectory.}

{Diffusion-ES and CEM-DP reach only $40.0\%$ and $33.3\%$ success, respectively, each incurring 16 obstacle collisions, consistent with the outer-loop search and single-candidate selection limitations discussed in 2D.} GRACE achieves the highest success rates at {$90.0\%$ for CNN and $93.3\%$ for UNet}, with only two and one obstacle collisions, respectively{, indicating that the advantages observed in 2D carry over to multiple simultaneous safety constraints. Its remaining failures arise when nearly all sampled rollouts violate the constraints, leaving no feasible sample to steer the guided mean away from collision, or when no feasible path to the goal is found within the step budget. With the lightweight CNN backbone, the total computation time is further reduced to $258$\,ms. PO-MPPI, CEM-DP, and GRACE require nearly identical guidance time of approximately $181$\,ms, so these differences arise under comparable sampling costs.}

{We also evaluated GRACE with different execution horizons $H_{\mathrm{exec}}$. GRACE (UNet) maintained the same $93.3\%$ success rate for $H_{\mathrm{exec}}\in\{8,16,32\}$. A larger $H_{\mathrm{exec}}$ reduces replanning frequency but executes a longer trajectory segment before incorporating the updated state. In our static setting, the environment and task constraints remain unchanged during execution, limiting the effect of this reduced adaptivity over the tested range.} Fig.~\ref{fig5} shows a representative GRACE (UNet) execution in which the manipulator reaches the target while satisfying the safety constraints in the cluttered environment.

\subsection{Real-World Experiments}
\label{subsec:real}
\begin{figure}[t]
    \centering
    \subfloat[]{%
        \includegraphics[width=0.589\linewidth]{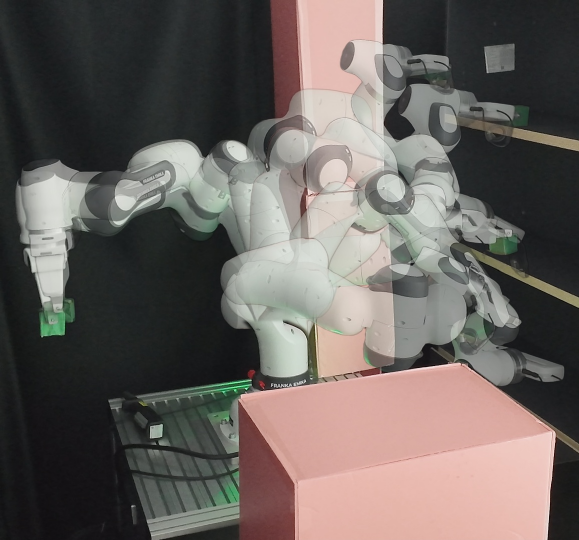}%
        \label{fig6a}%
    }%
    \hfill
    \begin{minipage}[b]{0.4\linewidth}
        \centering
        \subfloat[]{%
            \includegraphics[width=\linewidth]{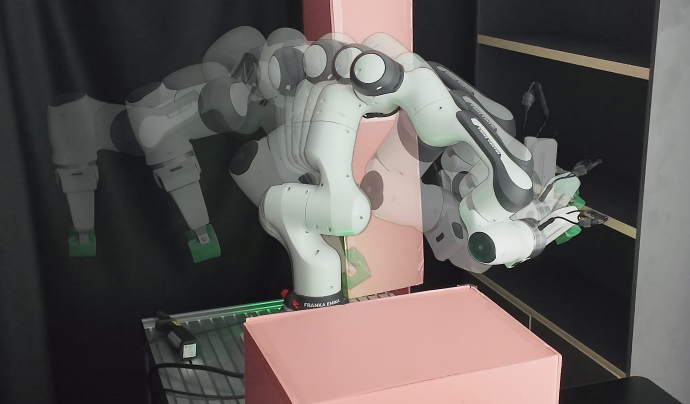}%
            \label{fig6b}%
        }\\
        \vspace{-2mm}
        \subfloat[]{%
            \includegraphics[width=\linewidth]{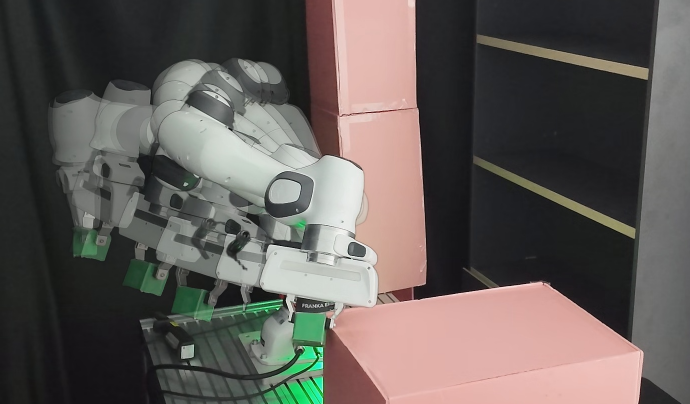}%
            \label{fig6c}%
        }
    \end{minipage}
    \caption{Real-world experiment. {(a) Experimental setup under the third obstacle condition, with both obstacles (pink boxes), goals on all three shelf levels, and laterally randomized start and goal configurations. (b) Successful motion of GRACE and (c) failed motion of the unguided diffusion prior under the same condition, with the bottom shelf as the target.}}
    \label{fig6}
    \vspace{-5mm}
\end{figure}

GRACE is validated on a physical 7-DoF FR3 manipulator that places a grasped object on a bookshelf while avoiding obstacles introduced near the shelf at deployment time. The diffusion prior is trained on demonstrations collected in a simulated replica of the workspace without these obstacles, and is deployed without retraining. The action is the joint configuration $\mathbf{q}$ with the gripper command, and the release is produced by the policy rather than triggered externally. {On hardware, GRACE uses $H=32$ and adaptively determines $H_{\mathrm{exec}}$ from the 32-step prediction. The next planning update is triggered before the current chunk is exhausted based on the recent inference time, and the new trajectory replaces the remainder once available. Each trial begins with the object already grasped and targets one of the three shelf levels, with the start configuration and goal placement randomly offset laterally. Three obstacle conditions are tested, with a low obstacle on the left of the shelf, a tall obstacle on the right, and both together. Fig.~\ref{fig6a} illustrates this setup.} The obstacle, self-collision, and workspace-bound costs are the only terms added at inference time.

{Five trials are run per shelf level under each condition, giving 45 trials each for GRACE (UNet) and the unguided diffusion prior (UNet). The unguided prior succeeds in 7, 10, and 0 of the 15 trials under the three conditions, respectively, only where a demonstrated mode happens to remain clear, and fails in every trial once both obstacles are present. GRACE succeeds in 14, 15, and 14 of 15, with no obstacle collision in any trial. The failure under the first condition is a placement inaccuracy, in which the manipulator clears the obstacle but misplaces the object on the shelf, and under the third the MPPI guidance falls into a local minimum between the combined obstacles. Across the three obstacle conditions, GRACE achieves 43/45 successful trials without obstacle collisions, substantially outperforming the unguided prior with 17/45 successes. Figs.~\ref{fig6b} and~\ref{fig6c} show a trial under the third condition with the bottom shelf as the target, in which GRACE detours around the obstacles and places the object, whereas the unguided prior follows a demonstrated mode blocked by the obstacles and fails.}

\section{CONCLUSION}
This paper presented GRACE, a gradient-free framework that integrates MPPI-based cost guidance into each diffusion reverse step by shifting the reverse mean while preserving the diffusion-schedule covariance, enabling diffusion policies to incorporate binary, discontinuous, and forward-evaluable deployment-time costs without retraining or backpropagated cost gradients. {GRACE achieved the highest success rates in 2D and 3D simulations while producing multiple distinct trajectory modes, and achieved 43/45 successes without obstacle collisions on hardware, compared with 17/45 for the unguided prior.} Future work will extend GRACE to broader robotic tasks and develop guidance that steers a single multi-task diffusion prior toward a desired task at inference time without training separate models.

\bibliographystyle{IEEEtran}
\bibliography{IEEEabrv, references}

\newpage

\vfill

\end{document}